\pdfoutput=1
\documentclass[11pt,letterpaper]{article}

\usepackage[T1]{fontenc}
\usepackage[sc]{mathpazo}
\usepackage[textwidth=5.6in,textheight=9.05in,top=1.0in,headsep=20pt,footskip=32pt]{geometry}
\usepackage{microtype}
\newcommand{\code}[1]{{\fontencoding{OT1}\fontfamily{cmtt}\selectfont #1}}
\usepackage{amsmath,amssymb}
\usepackage{booktabs}
\usepackage{xcolor}
\usepackage{tikz}
\usetikzlibrary{arrows.meta,calc}
\usepackage{pgfplots}
\pgfplotsset{compat=1.17}
\usepgfplotslibrary{groupplots}
\usepackage[font=small,labelfont=bf,labelsep=period,justification=justified,singlelinecheck=false,skip=6pt]{caption}
\usepackage{titlesec}
\usepackage{fancyhdr}
\usepackage[authoryear,round,semicolon]{natbib}
\usepackage[hidelinks,pdftitle={The Router Already Knows},
  pdfauthor={Andrew Bistras; Marty Johnson},
  pdfsubject={Native-router sparse dictionaries (NR-SAE) for mixture-of-experts models},
  pdfkeywords={NR-SAE, mixture-of-experts, sparse autoencoders, transcoders, mechanistic interpretability}]{hyperref}

\titleformat{\section}{\large\scshape}{\thesection}{0.9em}{}
\titlespacing*{\section}{0pt}{2.4ex plus .6ex minus .2ex}{1.0ex plus .2ex}
\titleformat{\subsection}{\normalsize\itshape}{\thesubsection}{0.8em}{}
\titlespacing*{\subsection}{0pt}{1.8ex plus .4ex minus .2ex}{0.5ex plus .1ex}

\fancypagestyle{plain}{\fancyhf{}\fancyfoot[C]{\small\thepage}}

\newcommand{\nr}{\mbox{NR-SAE}}
\newcommand{\para}[1]{\emph{#1}\enspace}

\begin{document}

\thispagestyle{plain}
\begin{center}
{\LARGE The Router Already Knows\par}
\vspace{1.2em}
{\scshape Andrew Bistras\textsuperscript{\,1,2}\qquad Marty Johnson\textsuperscript{\,1,2}\par}
\vspace{0.45em}
{\small\itshape \textsuperscript{1}University of California, Los Angeles\quad
\textsuperscript{2}Astrid Cybernetics\par}
\vspace{0.3em}
{\small Preprint, October 2026\par}
\end{center}
\vspace{0.6em}

\begin{quote}\small
\noindent\textsc{Abstract.}\enspace
Mixture-of-experts (MoE) models already make a sparse choice for every token: which experts
should process it. We ask whether a sparse dictionary should reuse that choice. \nr{} gives
each routed expert a small dictionary and opens only those the frozen router selects. At block
15 of Gemma~4 26B-A4B, after ten million C4 training tokens and across three seeds, the
expert-aligned autoencoder explains 0.518 of the variance in router-weighted expert outputs on a
fixed evaluation capture. A rank-addressed control with the same parameters but sixteen times as
many candidates reaches 0.481, a candidate-matched control 0.458, and an architecture-matched
control with randomized expert identity 0.387.
A transcoder variant also leads all three controls: 0.432, against 0.347, 0.371, and 0.118.
Part of this advantage is architectural: each expert writes into a subspace of at most a quarter
of the model width, so naming the expert names the subspace. How much it explains, we have not
measured. At this layer and budget, the router already knows something a dictionary would
otherwise have to learn. Our implementation did not
train faster, and semantic interpretability, downstream loss, and causal faithfulness remain
untested.
\end{quote}
\begin{center}\rule{0.9in}{0.4pt}\end{center}
\vspace{-0.4em}

\section{Introduction}

We propose that a sparse dictionary for a mixture-of-experts layer need not learn where to look.
The router already knows.

A mixture-of-experts (MoE) layer owns many parameters and consults few of them. A router selects
a handful of expert networks for each token, and their weighted outputs are summed
\citep{shazeer2017}. A sparse autoencoder (SAE) makes a second sparse choice, writing a dense
activation as a few directions from a much larger learned dictionary
\citep{cunningham2024,gao2025}. Where the dictionary sits decides what structure it can see.
Placed after the sum, it sees only the recombined update. Placed across the individual expert
outputs without their identities, it treats them as one population. Either way the router's
partition is discarded, unless someone hands it back.

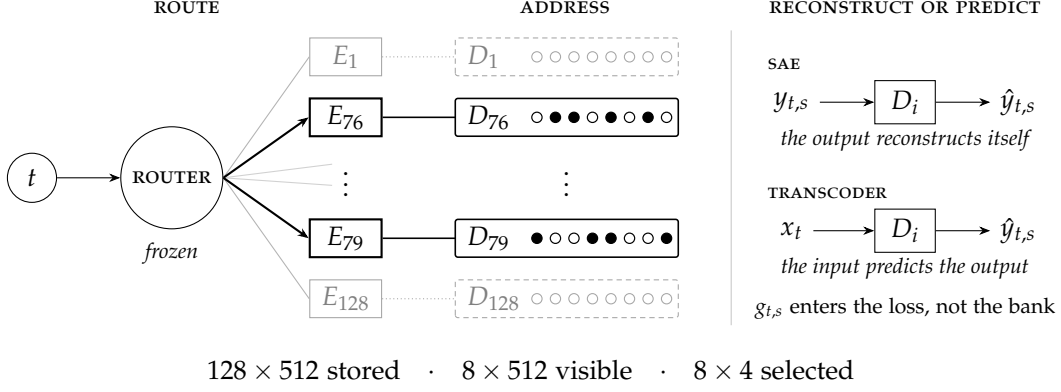
\begin{figure}[!h]
\centering
\begin{tikzpicture}[
  font=\small,
  >={Stealth[length=4.2pt,width=3.2pt]},
  exp/.style={draw, minimum width=9.5mm, minimum height=5mm, inner sep=1pt},
]
\node[font=\footnotesize\scshape] at (2.05,2.25) {route};
\node[font=\footnotesize\scshape] at (7.05,2.25) {address};
\node[font=\footnotesize\scshape] at (11.55,2.25) {reconstruct or predict};
\node[circle, draw, minimum size=6.5mm, inner sep=0pt] (t) at (0,0) {$t$};
\node[circle, draw, minimum size=13.5mm, inner sep=0pt, font=\footnotesize\scshape] (r) at (1.85,0) {router};
\node[font=\scriptsize\itshape, anchor=north] at (1.85,-0.72) {frozen};
\draw[->] (t) -- (r);
\node[exp, draw=black!35, text=black!50] (e1)   at (4.15, 1.6) {$E_{1}$};
\node[exp, line width=0.8pt]            (e37)  at (4.15, 0.8) {$E_{76}$};
\node                                    (ed)   at (4.15, 0.02) {$\vdots$};
\node[exp, line width=0.8pt]            (e94)  at (4.15,-0.8) {$E_{79}$};
\node[exp, draw=black!35, text=black!50] (e128) at (4.15,-1.6) {$E_{128}$};
\draw[black!30] (r.east) -- (e1.west);
\draw[black!30] (r.east) -- (e128.west);
\draw[black!18] (r.east) -- ($(ed.west)+(0,0.16)$);
\draw[black!18] (r.east) -- ($(ed.west)+(0,-0.12)$);
\draw[->, line width=0.8pt] (r.east) -- (e37.west);
\draw[->, line width=0.8pt] (r.east) -- (e94.west);
\newcommand{\bank}[4]{%
  \ifnum#3=1
    \draw[rounded corners=1pt, line width=0.6pt] (5.6,#1-0.25) rectangle (8.55,#1+0.25);
    \node[anchor=west] at (5.62,#1) {$D_{#2}$};
    \foreach \k in {0,...,7} {\draw[line width=0.45pt] (6.68+\k*0.245,#1) circle (0.068);}
    \foreach \k in {#4} {\fill (6.68+\k*0.245,#1) circle (0.068);}
  \else
    \draw[rounded corners=1pt, black!35, dash pattern=on 2pt off 1.5pt] (5.6,#1-0.25) rectangle (8.55,#1+0.25);
    \node[anchor=west, black!50] at (5.62,#1) {$D_{#2}$};
    \foreach \k in {0,...,7} {\draw[black!35, line width=0.4pt] (6.68+\k*0.245,#1) circle (0.068);}
  \fi}
\bank{1.6}{1}{0}{}
\bank{0.8}{76}{1}{1,2,4,6}
\node at (7.07,0.02) {$\vdots$};
\bank{-0.8}{79}{1}{0,3,4,7}
\bank{-1.6}{128}{0}{}
\draw[black!35, densely dotted] (e1.east) -- (5.6,1.6);
\draw[black!35, densely dotted] (e128.east) -- (5.6,-1.6);
\draw[line width=0.6pt] (e37.east) -- (5.6,0.8);
\draw[line width=0.6pt] (e94.east) -- (5.6,-0.8);
\draw[black!22] (9.25,-1.95) -- (9.25,1.95);
\node[font=\scriptsize\scshape, anchor=west] at (9.6,1.5) {sae};
\node (ys)  at (10.05,1.0) {$y_{t,s}$};
\node[draw, minimum width=8mm, minimum height=5mm] (ds) at (11.55,1.0) {$D_i$};
\node (yh1) at (13.05,1.0) {$\hat y_{t,s}$};
\draw[->] (ys) -- (ds); \draw[->] (ds) -- (yh1);
\node[font=\scriptsize\itshape] at (11.55,0.5) {the output reconstructs itself};
\node[font=\scriptsize\scshape, anchor=west] at (9.6,-0.2) {transcoder};
\node (xt)  at (10.05,-0.7) {$x_{t}$};
\node[draw, minimum width=8mm, minimum height=5mm] (dt) at (11.55,-0.7) {$D_i$};
\node (yh2) at (13.05,-0.7) {$\hat y_{t,s}$};
\draw[->] (xt) -- (dt); \draw[->] (dt) -- (yh2);
\node[font=\scriptsize\itshape] at (11.55,-1.2) {the input predicts the output};
\node[font=\scriptsize] at (11.55,-1.72) {$g_{t,s}$ enters the loss, not the bank};
\node[font=\small] at (6.6,-2.55) {$128\times512$ stored\quad$\cdot$\quad$8\times512$ visible\quad$\cdot$\quad$8\times4$ selected};
\end{tikzpicture}
\caption{\textbf{The router chooses the dictionary.} A pretrained MoE layer routes each token to
eight of its 128 experts (left). \nr{} reuses that selection to open the matching 512-feature
banks, each of which selects four features (center). A token therefore considers 4,096 of 65,536
stored features and selects 32, without a second router. The same bank layout is trained
separately for two tasks (right): reconstructing an expert's output, or predicting it from the
shared input. The shared expert is
unchanged and omitted.}
\label{fig:route}
\end{figure}

\nr{} hands it back. Each routed expert receives a small dictionary, and the frozen router serves
as its address: when expert 76 runs, dictionary 76 opens; otherwise dictionary 76 is neither
scored nor selected (Figure~\ref{fig:route}). Our question is narrower than whether MoE experts
are interpretable. It is whether the expert identities chosen by a pretrained router are useful
structure for sparse dictionary learning.

We contribute (i) the native-router factorization at a reconstructive and a predictive seam,
(ii) matched controls that separate stable expert identity from conditional capacity and
dictionary width, and (iii) a timing measurement showing that our grouped implementation does
not yet turn its arithmetic advantage into speed.

\section{Method}

\subsection{Expert-aligned dictionaries}

For token $t$, let $i_{t,s}$ be the expert at routing rank $s\in\{1,\dots,8\}$, $g_{t,s}$ its
router weight, and $y_{t,s}\in\mathbb{R}^{2816}$ its raw output, taken before routed summation
and normalization. Each of the 128 routed experts owns a bank of 512 features. Given an input
$u_{t,s}$, defined per seam in \S\ref{sec:seams}, the bank of the selected expert $i=i_{t,s}$
produces
\begin{equation}
\hat y_{t,s} \;=\; D_i\,\mathrm{TopK}_4\!\bigl(\mathrm{ReLU}(E_i\,u_{t,s}+b_i)\bigr).
\label{eq:bank}
\end{equation}
$\mathrm{TopK}_4$ keeps the four largest post-ReLU entries and zeroes the rest. Zero-valued
entries remain eligible, so four is a ceiling on the number of active features rather than a
promise. Centering, omitted from \eqref{eq:bank}, uses one pooled training mean of the targets,
and in the transcoder one of the inputs, shared by every bank in every layout; neither carries
expert identity. With eight routed experts per token, a
token sees 4,096 candidate features and selects 32 slots.

Training minimizes normalized squared error on router-weighted outputs over each batch
$\mathcal{B}$:
\begin{equation}
\mathcal{L}_{\mathcal{B}} \;=\;
\frac{\sum_{(t,s)\in\mathcal{B}}\,\lVert g_{t,s}(y_{t,s}-\hat y_{t,s})\rVert_2^2}
     {\sum_{(t,s)\in\mathcal{B}}\,\lVert g_{t,s}(y_{t,s}-\bar y)\rVert_2^2},
\label{eq:loss}
\end{equation}
where $\bar y$ is that pooled target mean. The denominator, the batch's total sum of squares
about $\bar y$, is the same for every layout in a panel, so losses are directly comparable, and
predicting $\bar y$ alone scores exactly zero in every layout. Router weights enter only here;
the banks receive unweighted inputs.

Part of what the router knows is plain arithmetic. Expert $i$ computes $y=W_i h$, where
$W_i\in\mathbb{R}^{2816\times704}$ is its down-projection and $h\in\mathbb{R}^{704}$ its hidden
activation. Its outputs therefore lie in a subspace of dimension at most 704, a quarter of the
model width. A dictionary that is told which expert ran is thereby told which subspace to
expect, before it has learned anything. Each 512-feature bank is smaller than that 704-dimension
ceiling, though we have not measured how many dimensions the outputs actually use.

\subsection{Two seams}\label{sec:seams}

We test the factorization at two points. The SAE receives the expert output itself,
$u_{t,s}=y_{t,s}$, and reconstructs it. The transcoder receives $x_t$, the common input to the
routed experts, and predicts each selected expert's output \citep{dunefsky2024}. The first asks
whether expert outputs admit expert-local dictionaries; the second, whether those dictionaries
can describe each expert's computation from its input.

\subsection{Relation to prior work}

Switch SAEs and multi-expert SAEs learn an auxiliary router over dense-model activations
\citep{mudide2025,xu2025}; the latter call such mixtures \mbox{MoE-SAEs}, which is one reason we do
not. RouterInterp runs in the opposite direction, using SAE features to explain a pretrained
MoE's routing, and proposes that each expert specializes in many fine-grained features rather
than one broad domain \citep{lasy2026}; per-expert dictionaries are well placed to test that
proposal. \citet{herbst2026} find that expert neurons are less polysemantic than dense
feed-forward neurons, and argue for interpreting experts directly, without a trained
dictionary. Their argument makes each expert's own neurons a natural reference point for our
banks, though not a matched one; we have not yet measured it (\S\ref{sec:limits}). Routers and
experts are also geometrically coupled during training \citep{ahrac2026}. Attention-output SAEs
train one dictionary on the concatenated outputs of all heads, before the output projection, so
features stay attributable to heads without splitting the dictionary \citep{kissane2024}. Mixtures of decoders are a recent
alternative to transcoders for sparse layer approximation \citep{oldfield2025}.

Here the native router is inherited and frozen, and its decisions index dictionaries trained on
the outputs it selected. The contribution is controlled evidence for that factorization, not a
new router.

\section{Experimental design}

\subsection{Matched controls}

A dictionary can look better because it stores more features, scores fewer candidates, or merely
preserves some partition. We compare four layouts (Table~\ref{tab:layouts}). All select 32 slots
per token and use the same loss.

\suppressfloats[t]
\begin{table}[t]
\centering\small
\caption{\textbf{Matched layouts.} All select 32 slots per token. F matches A's stored features
and parameters; B matches its candidates; R matches everything but a stable address, permuting
expert identities afresh for each sequence.}
\label{tab:layouts}
\begin{tabular}{@{}l l r r r@{}}
\toprule
Layout & Dictionary address & Banks $\times$ width & \begin{tabular}[b]{@{}r@{}}Stored\\features\end{tabular} & \begin{tabular}[b]{@{}r@{}}Candidates\\per token\end{tabular} \\
\midrule
A\enspace NR-SAE          & true expert identity & $128\times512$   & 65,536 & 4,096 \\
F\enspace Wide rank bank  & router rank, 1--8    & $8\times8{,}192$ & 65,536 & 65,536 \\
B\enspace Rank bank       & router rank, 1--8    & $8\times512$     & 4,096  & 4,096 \\
R\enspace Random identity & permuted identity    & $128\times512$   & 65,536 & 4,096 \\
\bottomrule
\end{tabular}
\end{table}

F, the wide rank bank, matches A's stored features and parameters but gives each of its eight
rank banks 8,192 candidates, sixteen times as many as an A bank. Of the two rank-addressed
layouts, it is the one matched to A in parameters.

B, the rank bank, matches A's candidate count with one 512-feature bank per routing rank, shared
across expert identities.

R keeps A's architecture exactly but applies a per-sequence bijection from expert IDs to bank
IDs, in training and evaluation alike. It preserves parameters, candidates, selected slots, the
routed components, and eight distinct banks per token, while destroying any stable relationship
between expert and bank. On average an R bank receives as much data as an A bank, but must
spend it on every expert's outputs rather than one expert's; in the transcoder, moreover, a bank
cannot tell which of the eight selected experts it is meant to predict.

No single control carries the claim; the three contrasts carry it together. A$-$F holds stored
capacity and parameters fixed, A$-$B the visible candidate budget, and A$-$R the architecture
itself, while removing the stable expert address. None of the four layouts is a global SAE over
the aggregated MoE output.

\subsection{Training and evaluation}

We train at block 15 (one-indexed) of the thirty in the pretrained Gemma~4 26B-A4B
\citep{gemma2026}, on packed C4 \citep{raffel2020}. Each run sees ten million tokens, 611 updates
of 16,384 tokens under one learning-rate schedule, with seeds 0--2. We evaluate checkpoints at
nominal 2M, 5M, and 10M tokens on one fixed evaluation capture: the same ordered 250,000 tokens
from the C4 validation split, in 625 packed sequences (\S\ref{sec:limits} explains why we do not
call it held out). All three checkpoints come from one cosine schedule, so the earlier two are snapshots
of an unfinished run rather than runs trained to those budgets.

The endpoint is component explained variance, $1-\mathrm{SSE}/\mathrm{SST}$, with one
denominator per panel. Intervals are 95\% paired hierarchical bootstrap intervals over seeds and
sequences, from 20,000 replicates, with no correction for multiplicity. With three seeds they
chiefly reflect variation across sequences; per-seed values are given in Table~\ref{tab:seeds}.

\begin{figure}[t]
\centering
\begin{tikzpicture}
\pgfplotsset{
  styA/.style={black, line width=1.1pt, mark=*, mark size=1.5pt},
  styF/.style={black!72, line width=0.7pt, dash pattern=on 4pt off 1.5pt on 1pt off 1.5pt,
               mark=o, mark size=1.6pt, mark options={solid, fill=white}},
  styB/.style={black!58, line width=0.7pt, dash pattern=on 3pt off 2pt,
               mark=square*, mark size=1.25pt, mark options={solid}},
  styR/.style={black!42, line width=0.7pt, densely dotted,
               mark=triangle*, mark size=1.8pt, mark options={solid}},
}
\tikzset{
  endlabA/.style={anchor=west, font=\footnotesize, text=black},
  endlabF/.style={anchor=west, font=\footnotesize, text=black!72},
  endlabB/.style={anchor=west, font=\footnotesize, text=black!58},
  endlabR/.style={anchor=west, font=\footnotesize, text=black!50},
}
\begin{groupplot}[
  group style={group size=2 by 1, horizontal sep=23mm},
  scale only axis, width=4.45cm, height=4.6cm,
  xmin=1.2, xmax=10.8, ymin=0, ymax=0.56,
  xtick={2,5,10}, xticklabels={2M,5M,10M},
  ytick={0,0.1,0.2,0.3,0.4,0.5},
  yticklabel style={/pgf/number format/fixed,/pgf/number format/fixed zerofill,/pgf/number format/precision=1},
  axis lines=left, axis line style={black!65, line width=0.45pt, -},
  tick style={black!65}, tick align=outside, major tick length=2.5pt,
  ymajorgrids, major grid style={black!10, line width=0.35pt},
  every tick label/.append style={font=\footnotesize},
  label style={font=\footnotesize},
  xlabel={training tokens},
  title style={font=\small\scshape, anchor=south west, at={(0,1.03)}},
  clip=false,
]
\nextgroupplot[title={Reconstruction (SAE)}, ylabel={component explained variance}]
\addplot[styR] coordinates {(2.0152,0.2624) (5.0135,0.3779) (10.0000,0.3873)};
\addplot[styB] coordinates {(2.0152,0.4069) (5.0135,0.4491) (10.0000,0.4576)};
\addplot[styF] coordinates {(2.0152,0.4176) (5.0135,0.4700) (10.0000,0.4809)};
\addplot[styA] coordinates {(2.0152,0.4452) (5.0135,0.5075) (10.0000,0.5184)};
\node[endlabA] at (axis cs:10.45,0.532) {A\enspace0.518};
\node[endlabF] at (axis cs:10.45,0.486) {F\enspace0.481};
\node[endlabB] at (axis cs:10.45,0.440) {B\enspace0.458};
\node[endlabR] at (axis cs:10.45,0.387) {R\enspace0.387};
\nextgroupplot[title={Prediction (transcoder)}]
\addplot[styR] coordinates {(2.0152,0.0116) (5.0135,0.1034) (10.0000,0.1183)};
\addplot[styB] coordinates {(2.0152,0.3168) (5.0135,0.3648) (10.0000,0.3715)};
\addplot[styF] coordinates {(2.0152,0.2757) (5.0135,0.3356) (10.0000,0.3468)};
\addplot[styA] coordinates {(2.0152,0.3579) (5.0135,0.4217) (10.0000,0.4317)};
\node[endlabA] at (axis cs:10.45,0.436) {A\enspace0.432};
\node[endlabF] at (axis cs:10.45,0.336) {F\enspace0.347};
\node[endlabB] at (axis cs:10.45,0.385) {B\enspace0.371};
\node[endlabR] at (axis cs:10.45,0.118) {R\enspace0.118};
\end{groupplot}
\end{tikzpicture}
\caption{\textbf{Expert identity leads throughout.} Three-seed mean component explained variance
on the fixed evaluation capture, at the three checkpoints of a single cosine schedule; the 2M and
5M points are therefore snapshots of unfinished runs. A leads every control at every checkpoint
in both panels. Per-seed values are in Table~\ref{tab:seeds}.}
\label{fig:curves}
\end{figure}
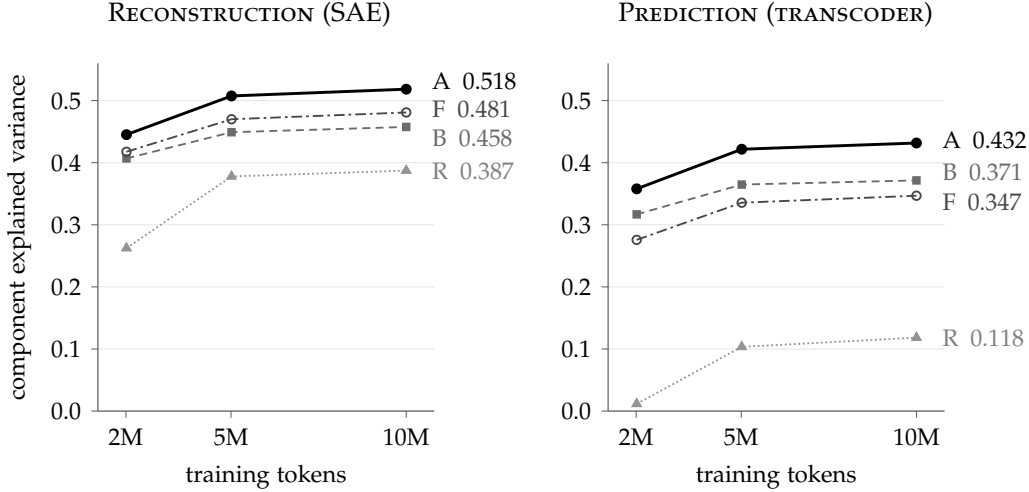

\section{Results}

\subsection{Component fidelity}

Table~\ref{tab:results} gives the endpoints and Figure~\ref{fig:curves} the learning curves. At
ten million tokens the aligned SAE explains 0.518 of component variance on the evaluation
capture. The wide rank bank F reaches 0.481, a margin of 0.038 [0.035,\,0.040]. The candidate-matched B
reaches 0.458, and R 0.387.

\begin{table}[t]
\centering\small
\caption{\textbf{Endpoints at ten million tokens.} Three-seed mean component explained variance
(EV) on the fixed evaluation capture. Margins are A minus the layout named, with 95\% paired
hierarchical bootstrap intervals; margins and intervals were computed before rounding.}
\label{tab:results}
\setlength{\tabcolsep}{5pt}
\begin{tabular}{@{}l cc cc@{}}
\toprule
 & \multicolumn{2}{c}{Reconstruction (SAE)} & \multicolumn{2}{c}{Prediction (transcoder)} \\
\cmidrule(lr){2-3}\cmidrule(l){4-5}
Layout & EV & A $-$ layout & EV & A $-$ layout \\
\midrule
A\enspace NR-SAE          & 0.518 & ---                   & 0.432 & --- \\
F\enspace Wide rank bank  & 0.481 & 0.038 [0.035, 0.040]  & 0.347 & 0.085 [0.082, 0.088] \\
B\enspace Rank bank       & 0.458 & 0.061 [0.059, 0.062]  & 0.371 & 0.060 [0.058, 0.062] \\
R\enspace Random identity & 0.387 & 0.131 [0.127, 0.135]  & 0.118 & 0.313 [0.308, 0.320] \\
\bottomrule
\end{tabular}
\end{table}

The lead survives the harder task of prediction. The aligned transcoder reaches 0.432, against
0.347 for F and 0.371 for B, margins of 0.085 and 0.060; the narrowest margin is thus against F
in reconstruction and against B in prediction. Against R the margin is 0.313. R withholds what
the router knows: each bank's data is spread across every expert and, in the transcoder, a bank
cannot tell which expert it is predicting. The A$-$R margins therefore measure that knowledge as a
whole, not inductive bias alone.

A leads every control at every checkpoint in both panels, and at every checkpoint its weakest
seed exceeds every control's strongest (Table~\ref{tab:seeds}). The plainest reading is that, at
this layer and budget, stable expert identity indexes a dictionary better than router rank,
random identity, or a much wider rank dictionary does. The pretrained router has supplied a
useful address.

\subsection{Systems cost}

A scores one sixteenth as many encoder candidates as F. Its decoding, dispatch, and optimizer
work are no smaller, and neither, so far, is its clock time. At ten million tokens A used 1.10--1.12 times F's training
chip-seconds in the SAE panel and 1.10--1.16 times in the transcoder panel, on same-generation
accelerators with matched timing boundaries (Appendix~\ref{app:repro}). We did not complete a
profile, so dispatch, the fixed-capacity scan, small matrix multiplications, and dense optimizer
work remain suspects rather than findings. A fused implementation is a separate project.

\section{Limitations and next steps}\label{sec:limits}

The result is evidence for conditional factorization: at this layer and budget, stable expert identities
are better dictionary addresses than router rank, random identity, or a much wider rank
dictionary.

\para{What comes free.}Each expert writes into a subspace of at most 704 of 2,816 dimensions
(\S2.1), so part of A's margin may be knowledge the router supplies before any dictionary is
trained. We have not measured how much. The obvious reference point is the expert's own neurons:
keep, for each routed expert, the four terms $h_j w_j$ of largest norm in $y=\sum_j h_j w_j$,
where $w_j$ is the $j$th column of $W_i$. This top-four native-neuron baseline needs no training,
but it is not a matched dictionary. It draws on 704 candidates rather than 512, its coefficients
are signed, and it reads the expert's hidden activations; in the transcoder it must run the very
expert it approximates. We have not run it. A learned router over the same 128 banks, in
the manner of Switch SAEs \citep{mudide2025}, would test whether native addresses are better than
learned ones or merely free.

\para{What is not shown.}Nothing here shows that individual features mean anything to a human.
We report component explained variance only, at a single sparsity: not routed-branch variance,
full-block reconstruction, logits, loss recovered, or any task endpoint. Fidelity after
recomposition is unknown. We have not shown semantic specialization, causal control, downstream
benefit, preserved perplexity, or faithful replacement of the MoE block. Partitioned
dictionaries can learn the same feature more than once \citep{xu2025}; we have not measured
whether ours do.

\para{Scope.}One model, one layer, one corpus, three seeds, and ten million tokens. At 611
updates the comparison measures early sample efficiency, not the asymptote. All layouts shared
one optimizer and schedule, and F, with the widest banks, may improve with its own tuning or a
longer horizon.

\para{Hygiene.}The evaluation capture was excluded from gradient updates but consulted during
earlier development of the activation rule, which is why we call it fixed rather than held out.
The six aligned runs were carried over from that earlier experiment, and their results were
known; before training the eighteen new controls we fixed the comparison matrix, the endpoint,
and the planned contrasts. Final evidence was content-hashed after it was generated, which
establishes lineage, not preregistration. Both asymmetries make the comparison less
confirmatory than it looks, and either could favor A. Retraining the aligned runs alongside the controls, and
evaluating on a fresh slice, would remove them.

\para{Next.}Replication across layers and a second open MoE would establish breadth. A sparsity
sweep with loss recovered would establish practical value. Feature labeling, intervention, and a
census of duplicated features would test whether the better dictionaries are also better
explanations. A fused sparse kernel would test whether the factorization can become a speedup.

\section{Conclusion}

At this layer and budget, native routing improves sparse component fidelity under controls for candidate count, parameter
count, selected-slot budget, and arbitrary partitioning. It does not yet make training cheaper,
and it has not yet been shown to make features clearer. The modest conclusion is enough: a
mixture-of-experts model has already carved its computation at useful joints, and a sparse
account of that computation should begin by respecting the cut.

Every routed model offers the same cut. We have examined one layer of one.

\vspace{0.9em}
\noindent\textsc{Acknowledgments.}\enspace This work was supported with Cloud TPUs from Google's
TPU Research Cloud (TRC). We thank the teams who made Gemma~4, C4, JAX, and MaxText available to
researchers. Claude (Anthropic) was used to edit the text, typeset the source, and redraw the figures
from our data. We have checked every claim and reference and take full responsibility for them.

\vspace{0.6em}
\noindent\textsc{Code and data.}\enspace Code, frozen configurations, aggregate results,
figures, and an animated routing schematic are available at
\href{https://github.com/andrewbistras/moe-sae}{\code{github.com/andrewbistras/moe-sae}}. Model
weights and source data remain subject to their original licenses.

\appendix
\renewcommand{\thetable}{A\arabic{table}}
\setcounter{table}{0}
\section{Reproducibility details}\label{app:repro}

\para{Runs.}The final comparison contains 24 training runs, six aligned runs carried over from
the preceding experiment and 18 newly trained controls, and 72 evaluated cells: two panels,
three seeds, four layouts, and three checkpoints. The nominal 2M and 5M checkpoints fall after
2,015,232 and 5,013,504 tokens (updates 123 and 306). The final batch masks its last 10,624
tokens, so training ends at exactly 10,000,000.

\begin{center}\small
\captionof{table}{\textbf{Per-seed values.} Component explained variance for each seed, in ascending order. At
every checkpoint in both panels, A's lowest seed exceeds every control's highest.}
\label{tab:seeds}
\setlength{\tabcolsep}{6pt}\renewcommand{\arraystretch}{0.94}
\begin{tabular}{@{}l ccc@{}}
\toprule
 & 2M & 5M & 10M \\
\midrule
\multicolumn{4}{@{}l}{\textit{Reconstruction (SAE)}}\\[1pt]
\quad A\enspace NR-SAE & 0.444\enspace 0.445\enspace 0.447 & 0.507\enspace 0.508\enspace 0.508 & 0.518\enspace 0.518\enspace 0.519 \\
\quad F\enspace Wide rank bank & 0.417\enspace 0.418\enspace 0.418 & 0.468\enspace 0.471\enspace 0.472 & 0.478\enspace 0.481\enspace 0.483 \\
\quad B\enspace Rank bank & 0.405\enspace 0.407\enspace 0.409 & 0.448\enspace 0.449\enspace 0.450 & 0.457\enspace 0.458\enspace 0.458 \\
\quad R\enspace Random identity & 0.262\enspace 0.262\enspace 0.263 & 0.378\enspace 0.378\enspace 0.378 & 0.387\enspace 0.387\enspace 0.388 \\
\addlinespace[4pt]
\multicolumn{4}{@{}l}{\textit{Prediction (transcoder)}}\\[1pt]
\quad A\enspace NR-SAE & 0.357\enspace 0.358\enspace 0.359 & 0.421\enspace 0.422\enspace 0.422 & 0.432\enspace 0.432\enspace 0.432 \\
\quad F\enspace Wide rank bank & 0.275\enspace 0.276\enspace 0.276 & 0.335\enspace 0.336\enspace 0.336 & 0.345\enspace 0.347\enspace 0.349 \\
\quad B\enspace Rank bank & 0.315\enspace 0.317\enspace 0.318 & 0.364\enspace 0.365\enspace 0.366 & 0.371\enspace 0.371\enspace 0.373 \\
\quad R\enspace Random identity & 0.011\enspace 0.011\enspace 0.013 & 0.101\enspace 0.104\enspace 0.105 & 0.117\enspace 0.118\enspace 0.120 \\
\bottomrule
\end{tabular}
\end{center}

\para{Optimization.}Parameters and matrix multiplications use bfloat16; Adam moments and
reductions use float32. Adam uses $\beta_1=0.9$, $\beta_2=0.999$, $\epsilon=10^{-8}$, no weight
decay, and global gradient clipping at 1.0. The learning rate warms up linearly over 61 updates
to $3\times10^{-4}$, then follows a cosine decay to $3\times10^{-5}$. Decoder columns are
initialized at unit norm and encoders at the decoder transpose, with zero encoder biases.
Decoder gradients are projected onto the tangent space, and columns are renormalized.

\para{Activation and dispatch.}ReLU followed by TopK, with zero-valued entries eligible for
selection; ties break by ascending global latent ID. Expert dispatch uses a 10\% capacity
margin, and any capacity overflow or dropped token is a hard error.

\para{Timing scope.}The timed region includes the fixed-capacity scan and the synchronized
optimizer loop, including time spent waiting for input. It excludes prefetch; one-time
accelerator and executable-graph preflight; lowering and compilation; checkpoint serialization;
and evaluation.

\para{Model and data.}The model is \code{google/gemma-4-26B-A4B}; the data are
\code{allenai/c4}, configuration \code{en}, with the training split for optimization and the
validation split for evaluation. Revisions, and the canonical identities computed after the
results existed (these are not raw-file digests):
\par\smallskip
{\centering\footnotesize
\begin{tabular}{@{}l l@{}}
model revision   & \code{24548b62aa021d562695c04aaf7758a1ea47990b}\\
data revision    & \code{1588ec454efa1a09f29cd18ddd04fe05fc8653a2}\\
report identity   & \code{fc1d6d4112dbd2f5cea075e1c257424c7fc6e917d66b9e6c5d42814a0940f460}\\
evidence identity & \code{4b8bf760befc6b8dbca2efe209e5de9c1ed47ce67f46e3a1cd9e0dbc9370329f}\\
\end{tabular}\par}

\end{document}